# Development of an Improved Capsule-Yolo Network for Automatic Tomato Plant Disease Early Detection and Diagnosis

[1]Idris Ochijenu, [2]Monday Abutu Idakwo, [3]Sani Felix
[1]Computer Science Department, Salem University, Lokoja, Kogi State.
[2]Department of Computer Engineering,
Faculty of Engineering, Federal University Lokoja, Kogi State.,
[3]Computer Science Department, Federal Polytechnic Idah, Kogi State

**ABSTRACT**

*Like many countries, Nigeria is naturally endowed with fertile agricultural soil that supports large-scale tomato production. However, the prevalence of disease-causing pathogens poses a significant threat to tomato health, often leading to reduced yields and, in severe cases, the extinction of certain species. These diseases jeopardize both the quality and quantity of tomato harvests, contributing to food insecurity. Fortunately, tomato diseases can often be visually identified through distinct forms, appearances, or textures—typically first visible on leaves and fruits. This study presents an enhanced Capsule-YOLO network architecture designed to automatically segment overlapping and occluded tomato leaf images from complex backgrounds using the YOLO framework. It identifies disease symptoms with impressive performance metrics: 99.31% accuracy, 98.78% recall, and 99.09% precision, and a 98.93% F1-score— representing improvements of 2.91%, 1.84%, 5.64%, and 4.12% over existing state-of-the-art methods. Additionally, a user-friendly interface was developed to allow farmers and users to upload images of affected tomato plants and detect early disease symptoms. The system also provides recommendations for appropriate diagnosis and treatment. The effectiveness of this approach promises significant benefits for the agricultural sector by enhancing crop yields and strengthening food security.*

## INTRODUCTION

Tomatoes are nutrient-rich vegetables that play a crucial role in human health, owing to their essential nutrients like folate, vitamin C, and potassium. Their low calories make them ideal for weight management and healthy diet promotion. Beyond their nutritional values, tomatoes are versatile ingredients used in a variety of sauces and condiments to improve food flavor (Idakwo et al., 2024). From an economic perspective, tomato cultivation significantly contributes to the agricultural industry. For instance, the 2020 Nigerian vegetable output revealed that 15.7 million tons, which includes 3.7, 1.8, 1.4, 1.0, and 0.7 million tons of tomatoes, okra, onions, cucumber, and ginger, were produced respectively (Ofuya et al., 2023). The sale and production of tomatoes stimulate various related sectors, including seed production, retail, transportation, and greenhouse construction. Additionally, tomatoes positively impact the ecosystem through the absorption of substantial amounts of carbon dioxide, which assists in the mitigation of greenhouse gas emissions, provides habitats for biodiversity, and supports ecological balance. Tomatoes also hold substantial research value, serving as model organisms in biotechnology, genetics, molecular biology, cell biology, and genomics (Zayani et al., 2024).

*Corresponding author: Monday Abutu Idakwo*
✉ monday.idakwo@fulokoja.edu.ng
*Department of Computer Engineering, Faculty of Engineering, Federal University Lokoja, Kogi State.*

Nonetheless, the rapid climatic changes and increase in disease-causing pathogens have resulted in lower yields and created major losses to agriculture and the economy. This spurt in novel diseases has necessitated the need for earlier identification with necessary action to avert the consequences. Physical and access constraints impede early diagnosis, leading to inadequate treatment; thus, an autonomous computational system to assist the farmers becomes necessary. Several developed computer vision applications fulfill this need as they analyze and process crop leaf images using image processing and machine/deep learning methods (Li et al., 2020).

Disease symptoms are commonly exhibited in leaves first and, therefore, serve as an early source for disease identification (Arivazhagan et al., 2013). Different diseases can show varying visual features in crop leaves, and analyzing these features through images using deep learning or machine learning algorithms helps identify and classify diseases precisely (Joseph et al., 2023).

Deep learning algorithms have stirred a paradigm shift in plant disease detection ever since their evolution, especially in identifying leaf disease for effective diagnosis. This evolution, driven by computer vision approaches and neural networks, has contributed to precision agriculture progress and led to a wider creation of improved deep learning and machine learning models tailored to accurately diagnose leaf diseases, including tomatoes. These models analyze leaf images and classify diseases according to their visual symptoms. Despite the success of deep learning methods, their larger parameter numbers, complex network structures, computational complexity, and higher GPU hardware demands are the inherent limitations that need to be addressed. Interestingly, the Capsule network has evolved to solve the larger data requirements, complex networks, and computational complexity of the convolutional neural network.

Furthermore, capsule networks resolve the convolutional neural network limitation in recognising object pose and deformation. Notwithstanding, classical capsule networks are computationally expensive during the training phase of the dataset (Yoro et al., 2024), despite having the shortest prediction time compared to other deep-learning approaches, as demonstrated by researchers (Idakwo et al., 2024). Therefore, this paper integrated the You Only Look Once (YOLO) (Zayani et al., 2024) fast detection approach into the Capsule network to produce an improved capsule-YOLO network with lower computational demand, faster extraction of disease symptoms under controlled and uncontrolled environments.

## LITERATURE REVIEW

Tomato leaf segmentation and disease symptom detection have remained an active research area with diverse techniques reported in the literature. Idakwo et al., (2024a) proposed an improved tomato ripeness detection and sorting system that utilizes the equivariance property of capsule network to effectively classify tomatoes into their respective ripeness stages with an average performance of 99.56%, 96.20%, 96.20%, and 96.40% qs accuracy, precision, recall, and F1-Score respectively. The inherent capsule network poses parameters that eliminate the need for data augmentation strategies. Thus, allowing fewer images to be trained, reducing computation time.

Zayani et al. (2024) explored the inherent YOLOv8 features and developed an automated tomato disease detection system. The developed system augmented the Roboflow dataset and achieved an overall accuracy of 66.67%. Nonetheless, the data imbalance nature led to a class-specific performance, highlighting challenges in differentiating certain diseases. Therefore, data balancing, exploring alternative architectures, and adopting disease-specific metrics. Joshi et al. (2025) introduced a hybrid data augmentation approach and increased the tomato disease dataset size from 737 images to 6696 images, which improves the model accuracy and robustness. The YOLOv8n deep convolutional neural network was employed and detected seven different tomato diseases with 96.5 % mAP, 97 % precision, 93.8 % recall, and 95 % F1 score.

Corresponding author: Monday Abutu Idakwo
✉ monday.idakwo@fulokoja.edu.ng
Department of Computer Engineering, Faculty of Engineering, Federal University Lokoja, Kogi State.

From the literature reviewed, it is evident that the learning ability, flexibility, and adaptability of deep learning in detecting tomato disease symptoms while ensuring higher accuracy and faster identification of symptoms have been widely explored. Although some of the systems showed improved accuracy, the complex factors inherent in the effective detection of diseases under complex backgrounds, varying lighting conditions, image pose (orientation and size), and scaling above seven symptoms detection are the basis on which this paper is inspired. Furthermore, it is evident that no singular method has resolved all issues highlighted, and there is value in combining various factors to build and improve versions of these detection systems.

**YOLO Networks**

You Only Look Once introduced in 2015 by Joseph Redmon. has evolved over the years with variant version and revised edited (Wang & Liu, 2021). The YOLO object detection algorithm is renowned for its compact model size and fast computation speed. Its architecture is relatively simple, allowing the neural network to directly predict both the location and class of detected objects. One of YOLO's key advantages is its speed as it processes an image through the network just once, it can instantly generate detection results, making it well-suited for real-time analysis. YOLO operates on the entire image, enabling it to capture global context and minimize false detections caused by background confusion. It also demonstrates strong generalization capabilities by learning robust features that can be adapted across different domains.

By framing object detection as a regression task, YOLO streamlines the detection process. However, its accuracy still has room for improvement. The model tends to struggle with detecting densely packed or overlapping objects. This limitation arises because each grid cell predicts only two bounding boxes and assigns them to a single class, which can lead to incorrect aspect ratios and hinder performance in complex scenes. Due to the loss function, the positioning error is the main reason for improving the detection efficiency. Especially, the handling of large and small objects needs to be strengthened. In the implementation, the most important thing is how to design the loss function so that these three aspects can be well balanced. YOLO incorporates several down-sampling layers, which can lead to incomplete feature extraction of targets, potentially affecting detection accuracy. nevertheless, these features still contribute to enhancing the overall detection performance.

The original YOLO model is composed of 24 convolutional layers followed by two fully connected layers. It generates multiple bounding box predictions for each grid cell, and among them, the boxes with the highest Intersection over Union (IOU) compared to the ground truth are retained using a technique called non-maximum suppression (Wang & Liu, 2021). Despite its advantages, YOLO faces two main limitations: imprecise localization of objects and a lower recall rate compared to region proposal-based methods.

**Capsule Networks**

The Capsule Network introduced by Sabour et al. (2017) offers a novel method of image processing as capsule layers are incorporated to encode both the existence of a feature and spatial information (pose) within an image. This additional information can be particularly beneficial for segmenting overlapping leaves, where accurate delineation of boundaries between individual leaves is critical. The units in the capsule are linked through a dynamic routing mechanism as an alternative to the common down-sampling algorithms (such as max-pooling). Different modifications have been implemented to the original capsule network structure, and the various models designed to achieve higher performance in image classifications (Idakwo *et al.*, 2024a).

The capsule network creates a hierarchical representation of the image by passing it through layers just like a CNN. A capsule network uses two layers referred to as the primary and secondary capsule layers, unlike CNN, with many layers (Patrick *et al.*, 2021), as shown in Figure 1.

*Corresponding author: Monday Abutu Idakwo*
✉ *monday.idakwo@fulokoja.edu.ng*
*Department of Computer Engineering, Faculty of Engineering, Federal University Lokoja, Kogi State.*

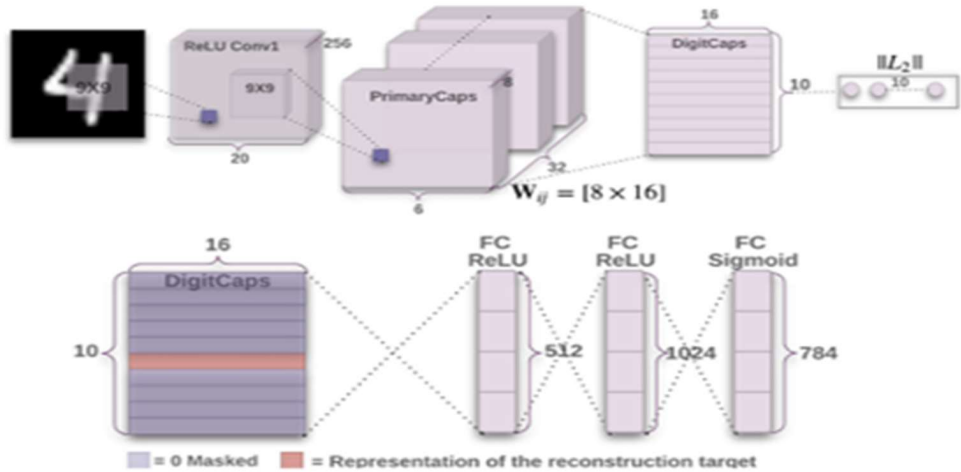

Figure 1: Structure of Capsule Network (Sabour et al. 2017)

The images' low-level features are captured in the primary layer. The secondary layer can predict the existence of an object and its pose information in the image. The main dissimilarity between the capsule network and the convolutional neural network is as summarized (Sabour et al., 2017).

### Performance Evaluation Metrics

The goal of developing any tomato disease detection system is for an early and accurate detection of the tomato disease symptoms before they cause considerable damage. Therefore, it becomes paramount to evaluate the system using the evaluation metrics discussed in this subsection.

### Accuracy

Accuracy is an intuitive measurement that shows imbalanced data. It is evaluated as the number of tomato diseases correctly identified using (Idakwo et al., 2024a):

$$Acc = (TP+TN)/(TP+TN+FP+FN) \qquad (1)$$

Where: TP represents True Positive, and is the data instances correctly Predicted by the classifier.

FN represents False Negative, and It is the data instances wrongly Predicted as Normal instances.

FP represents False Positive, and it is the data instances wrong classified.

TN represents True Negative, and It is the instance correctly classified as Normal instances.

### Precision

Precision evaluates the set of the correct ratio of correctly identified tomato diseases to all the samples predicted. It is mathematically given by (Yoro et al., 2024)

$$Pre = \frac{TP}{TP+FP} \times 100 \qquad (2)$$

### Recall

A recall is the ratio of all samples correctly classified to all the samples. It is also known as the detection rate and is given by (Idakwo et al., 2024a):

$$Pre = \frac{TP}{TP+FN} \times 100 \qquad (3)$$

### False Alarm Rate

The false alarm rate is also referred to as the false positive rate. It is defined as the ratio of wrongly predicted samples to all the normal samples. The false alarm rate is mathematically expressed as given in (Yoro et al., 2024).

$$FAR = \frac{FP}{FP+TN} \times 100 \qquad (4)$$

*Corresponding author: Monday Abutu Idakwo*
✉ monday.idakwo@fulokoja.edu.ng
*Department of Computer Engineering, Faculty of Engineering, Federal University Lokoja, Kogi State.*

## METHODOLOGY

Tomato leaves' shape, fruit, size, texture, colour, and other leaf characteristics can reveal signs of abiotic (nonliving) or biotic (living) stress factors impacting plant growth as discussed in chapter one. While some symptoms are minute, others are large and visible. Therefore, to improve the detection rate, this developed a hybrid framework that combines the strengths of Capsule Networks (CapsNets) and the You Only Look Once object detection system. The CapsNets-YOLO architecture consists of a convolutional backbone, a capsule-based feature encoding module, and a YOLO-inspired detection head.

The convolutional backbone extracts low-level and mid-level features from the tomato leaf image. These features are then fed into the capsule-based module, which replaces traditional pooling layers with dynamic routing mechanisms to preserve spatial hierarchies and part-whole relationships. In this module, capsules (groups of neurons) encode spatial information such as orientation, pose, and scale, allowing the network to handle the differences in tomato symptoms' occlusion. The routing mechanism uses an iterative agreement process that allows lower-level capsules to predict the output of higher-level capsules. The coupling coefficients are updated based on the degree of agreement between predictions. This enables only the propagation of relevant features are enhancing the network's ability to detect complex and overlapping objects. The output of the capsule module is then passed to the YOLO detection head, which predicts bounding boxes, class probabilities, and objectness scores. The YOLO head is modified to incorporate the rich spatial information provided by the capsules, improving localization accuracy and reducing false positives. The entire architecture was trained end-to-end using a composite loss function that combines localization loss, capsule reconstruction loss, and classification loss. The reconstruction loss, evaluated by decoding the capsule outputs and comparing them to the original input, acts as a regularizer, encouraging the network to learn more generalizable and robust features. The developed CapsNets-YOLO architecture has the added advantage of detecting objects in scenarios involving occluded objects, complex backgrounds, and varying viewpoints.

### Dataset

The dataset is an essential aspect needed for training any machine model. Therefore, tomato disease images in the benchmark PlantVillage dataset (Boukhris *et al.,* 2024) and PlantDoc dataset (Singh *et al.,* 2020) were extracted. While the PlantVillage dataset has ten tomato classes, the images have uniform background, relative intensity, and high fidelity, which is ideal for a controlled environment. The PlanDoc has seven tomato disease classes, but the images have varying quality (low quality, hi-fidelity, multiple backgrounds), which conform to a real farm scenario. Therefore, this paper combined both the controlled and uncontrolled environment features to produce an improved dataset of only tomato diseases amidst the other plant diseases in the benchmark datasets. Plant symptoms can be informed by local discoloration of the leaf, little spots or dots with a complex background, occlusion, variations in illumination, and leaf pose (Gong & Zhang, 2023). Therefore, it is essential to cover all these scenarios to increase the plant symptom detection rate. The summary of the dataset is shown in Table 1.

Table 1: Proposed Dataset Summary

| S/N | Diseases | PlantVillage Images | PlantDoc Images | PlantDocVill Images |
|-----|----------|---------------------|-----------------|---------------------|
| 1 | Bacterial Spot | 2,127 | 100 | 200 |
| 2 | Early Blight | 1,000 | 100 | 200 |
| 3 | Late Blight | 1,909 | 100 | 200 |
| 4 | Leaf Mold | 952 | 50 | 100 |

*Corresponding author: Monday Abutu Idakwo*
✉ *monday.idakwo@fulokoja.edu.ng*
*Department of Computer Engineering, Faculty of Engineering, Federal University Lokoja, Kogi State.*

| S/N | Diseases | PlantVillage Images | PlantDoc Images | PlantDocVill Images |
|---|---|---|---|---|
| 5 | Septoria Leaf Spot | 1,771 | 100 | 200 |
| 6 | Spider Mites | 1,676 | | 200 |
| 7 | Target Spot | 1,404 | 100 | 200 |
| 8 | Tomato Yellow Leaf Curl Virus | 5,357 | 50 | 100 |
| 9 | Tomato Mosaic Virus | 373 | | 200 |
| 10 | Healthy | 1,591 | | 200 |
| | Total | 18, 160 | 600 | 1,800 |

The PlantVillage dataset is known for its uniform background, relative intensity, and high-fidelity images. These ideal characteristics may not conform to natural agriculture farm images, whose image quality is dependent on weather conditions, pixel quality of the camera, background and overlapping or occluded images. Hence, the need to integrate the PlantDoc images with varying image quality (low quality, hi-fidelity, multiple backgrounds). However, direct integration of similar image symptoms will lead to a data imbalance problem. The seven common images were combined using the PlantDoc image number as the threshold.

Furthermore, to avoid the three remaining classes dominating the dataset, the maximum number from the combination of the two datasets was adopted as the threshold, as shown in Table 3.2. Therefore, the new dataset PlantDocVill has a total of 1,800 images with 7 classes of images showing varying characteristics like control environment, complex background, varying lighting conditions, low and high-fidelity images, and occluded images. The images were annotated with their respective disease labels and organized hierarchically based on disease categories, severity levels, and plant parts affected. This hierarchical structure was pivotal for enabling multi-level analysis and model training.

To create the HDF5 dataset, the preprocessed and annotated images were converted into a structured format that supports efficient storage and retrieval using the Python h5 library. Metadata, including hierarchical categories and image labels, were embedded within the HDF5 file to facilitate easy access and interpretation. The dataset was then balanced by ensuring an equal distribution of images across all disease categories and hierarchical levels, minimizing bias when subjected to the CapsNetYOLO architecture. Finally, the dataset was validated through statistical analysis and visual inspection to confirm its quality, balance, and adherence to the hierarchical structure. This comprehensive methodology ensures the creation of a robust, balanced, and well-organized HDF5 dataset suitable for training and evaluating machine learning models for tomato disease detection and classification.

## RESULTS AND DISCUSSION

This section discusses the results obtained from the implementation of the developed system. This entails the evaluation of the developed system's performance, comparative analysis with existing systems, and the percentage improvement. As discussed, the dataset is pivotal in the development of any machine learning model. The standard PlantVillage, PlantDoc dataset, and the balanced PlantDocVill dataset are presented in Figure 2.

*Corresponding author: Monday Abutu Idakwo*
✉ *monday.idakwo@fulokoja.edu.ng*
*Department of Computer Engineering, Faculty of Engineering, Federal University Lokoja, Kogi State.*

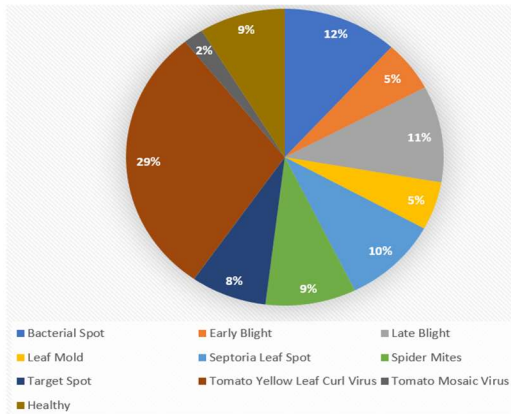

a. **PlantVillage Dataset**

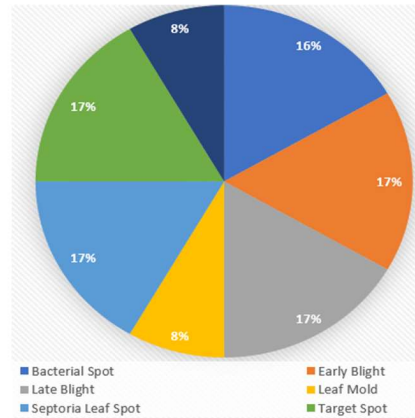

b. **PlantDoc Dataset**

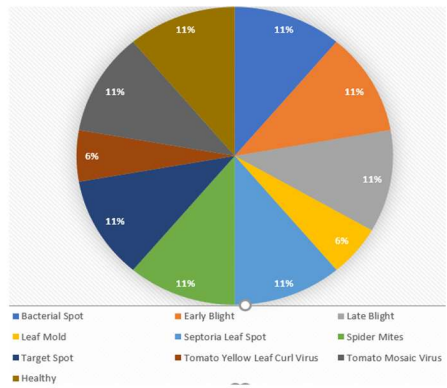

(c) **PlantDocVill Dataset**

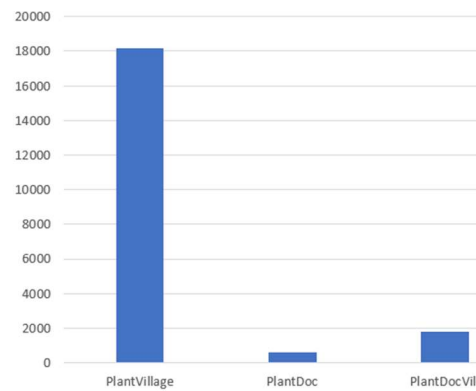

(d) **Comparison of the Dataset**

Figure 2: Improved PlantDocVill Dataset from the Benchmark Dataset

From Figure 2 (a), the PlantVillage dataset, well known for its uniform background, relative intensity, and high-fidelity images, contains a total tomatoes images of 18000 whose disease classes are not evenly distributed. While these images were obtained under a controlled environment, their higher numbers of classes (10) make them ideal for selection for any system whose aim is to improve on a larger range of disease detection. The PlantDoc, on the other hand, conformed with the conventional agricultural farming environment owing to their varying image quality (low quality, hi-fidelity, multiple backgrounds) but however, their fewer classes (7) created the need for a refined dataset as presented in Figure 2 (c). Unlike the benchmark images with wider disparity of image classes, the proposed PlanDocVill, as seen in Figure 2 (c), has a balanced dataset

The developed Improved CapsNet-YOLO Architecture was deployed to train the enhanced PlantDocVill Dataset, which is specifically designed to support plant disease classification. The dataset was partitioned into training and testing subsets, with 80% allocated for training and 20% reserved for testing. This ratio ensures that the model is exposed to a large portion of the data during training, while still being evaluated on a diverse and representative set of unseen data for robust performance evaluation.

*Corresponding author: Monday Abutu Idakwo*
✉ monday.idakwo@fulokoja.edu.ng
*Department of Computer Engineering, Faculty of Engineering, Federal University Lokoja, Kogi State.*

To optimize model convergence and prevent overfitting, the following hyperparameters were chosen:

1. Learning Rate (0.0001): A low learning rate of 0.0001 was selected to allow for gradual optimization of model weights. This small learning rate ensures the model trains efficiently without overshooting optimal values, which could destabilize the learning process.

2. Early Stopping: The early stopping mechanism was implemented to halt training if no improvement in validation loss is observed over a set number of epochs. This technique is crucial in preventing overfitting and reducing unnecessary computational cost, ensuring that the model stops training once it has reached optimal performance.

3. Epochs (40): A total of 40 epochs were used for training. This number was selected to provide sufficient time for the model to learn the intricate patterns of the dataset while avoiding excessive training that could lead to overfitting. This balance ensures that the model can generalize well to unseen data. The accuracy and loss of the model during the training and validation, showcasing the model's performance and the effectiveness of the chosen parameters, are presented in Figure 3.

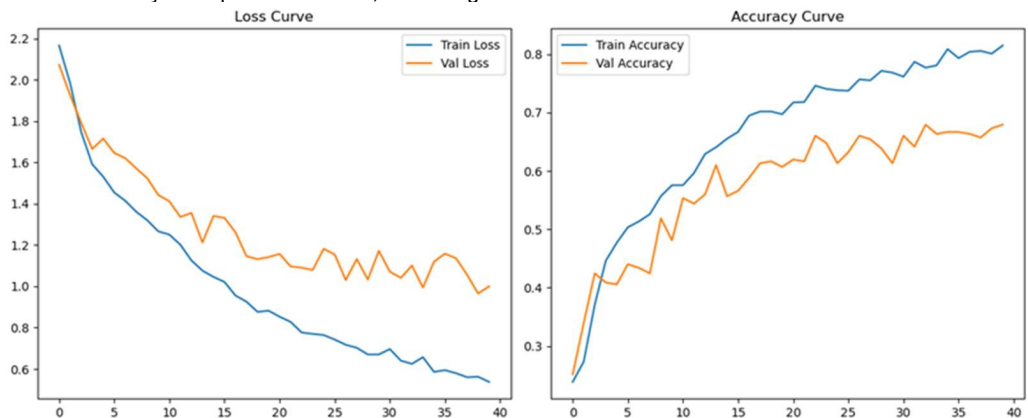

Figure 3: Accuracy and Loss During the Training and Validation of the Model

From Figure 3, the system was able to achieve a high accuracy during the training phase. Hence, the parameters contributed to the effectiveness and efficiency of the improved CapsNet-YOLO architecture, ensuring the model's generalizability, stability, and performance on the PlantDocVill dataset.

Additionally, the confusion matrix was used to evaluate the performance of the developed model. The result obtained is presented in Figure 3.

*Corresponding author: Monday Abutu Idakwo*
✉ *monday.idakwo@fulokoja.edu.ng*
*Department of Computer Engineering, Faculty of Engineering, Federal University Lokoja, Kogi State.*

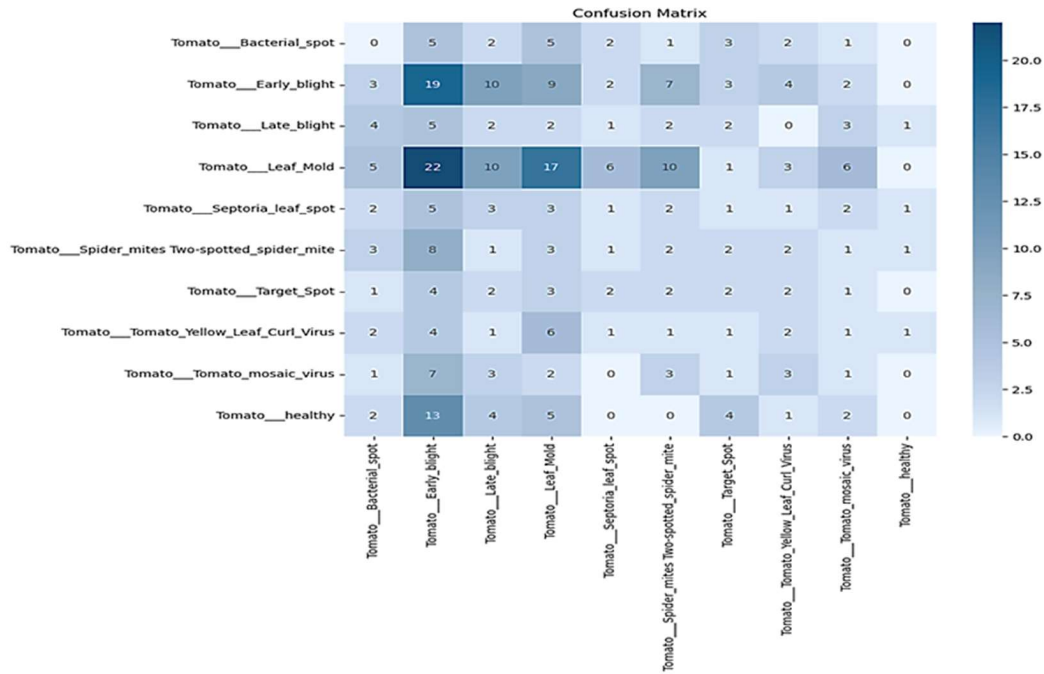

Figure 3: Model Confusion Matrix

The obtained confusion matrix provided in Figure 3 evaluates the performance of a tomato plant disease classification model by comparing predicted labels against actual disease categories. Each row represents a true disease class (such as Bacterial Spot, Early Blight, or healthy plants), while the numbers in the rows indicate how often the model correctly or incorrectly classified those instances. The diagonal values highlight correct predictions, while off-diagonal numbers reveal misclassifications. The matrix shows uneven performance, with some diseases like Leaf Mold achieving relatively higher correct predictions (22 instances) but still suffering from significant misclassifications. In contrast, diseases like Late Blight perform poorly, with only 2 correct identifications. The model struggles to distinguish between certain disease pairs, indicating a need for improved training data or model adjustments to reduce confusion.. Overall, the matrix identifies key areas for improvement, particularly in enhancing accuracy for frequently confused diseases and reducing false positives.

Furthermore, to evaluate the performance of the system on unseen tomato disease images, a user-friendly interface was developed and utilized, as depicted in Figure 4. This interface was designed to facilitate the easy input and processing of new tomato disease images, allowing users to seamlessly interact with the model without requiring technical expertise. Hence, the interface serves as a crucial tool for real-world deployment, enabling users to upload images and obtain real-time predictions as presented in Figure 4.

*Corresponding author: Monday Abutu Idakwo*
✉ monday.idakwo@fulokoja.edu.ng
*Department of Computer Engineering, Faculty of Engineering, Federal University Lokoja, Kogi State.*

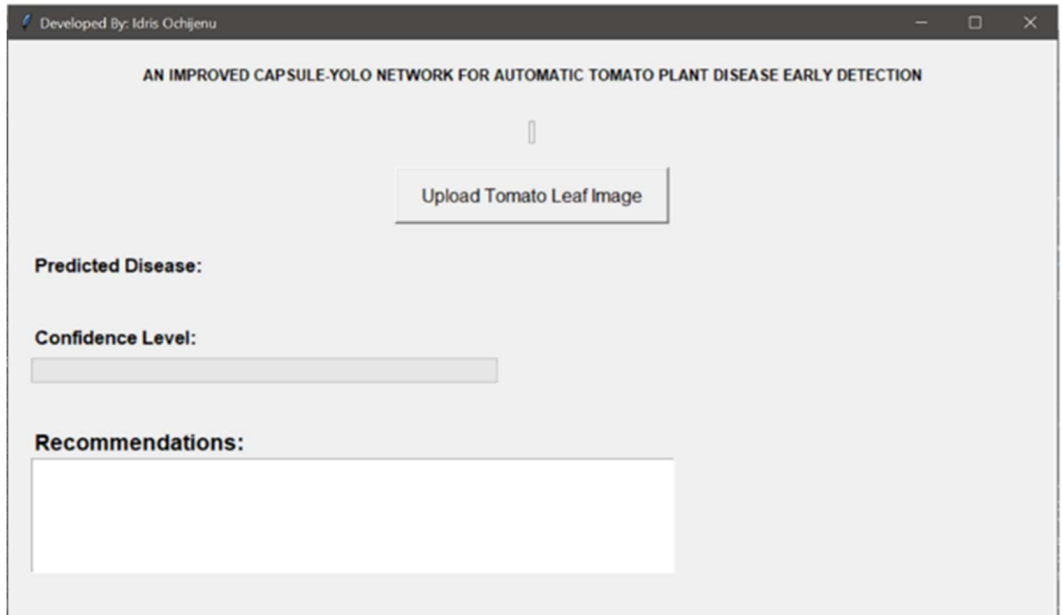

Figure 4: Developed System User-friendly Interface

From Figure 4, the developed user interface allows the user to upload a tomato leaf image for the model to predict the disease based on the visible symptoms. The confidence level is an indicator to show the extent to which the model is certain about the prediction. This will equally guide the model in the expected recommendations that will guide the farmers in treating the disease. To visualize this, a reserved tomato image from discarded images in the PlantVillage and PlantDoc datasets was randomly selected and subjected to prediction. The results obtained are presented in Figure 5.

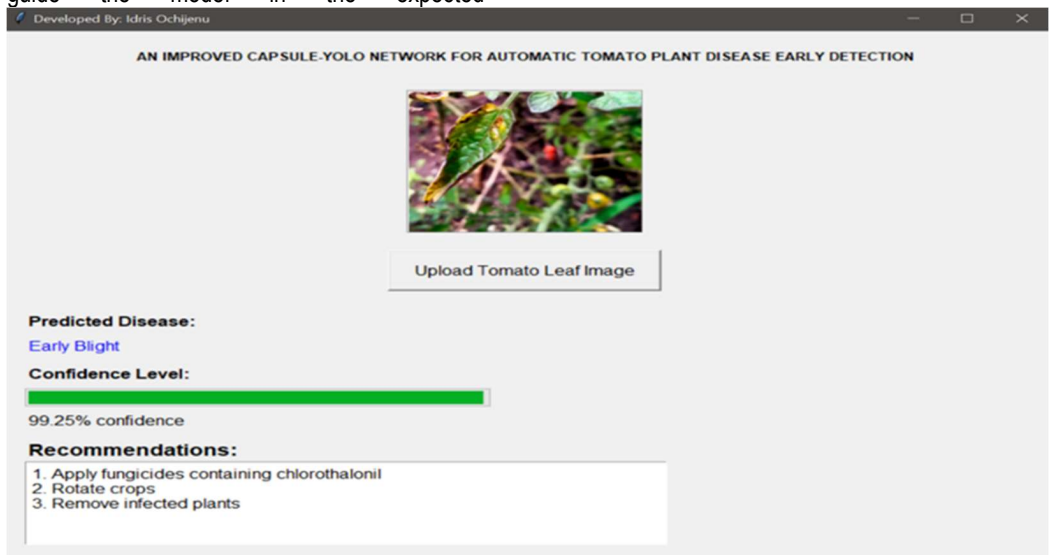

Figure 5: Disease Predictions and Recommendations

Corresponding author: Monday Abutu Idakwo
✉ monday.idakwo@fulokoja.edu.ng
Department of Computer Engineering, Faculty of Engineering, Federal University Lokoja, Kogi State.

As shown in Figure 5, the developed system successfully identified the disease in the uploaded image as Early Blight, despite the complex background of the tomato plant image. The system achieved a high confidence level of 99.25%, demonstrating its robust ability to accurately detect and classify diseases even in challenging conditions.

Additionally, the system provides valuable support to users or farmers by offering a detailed treatment recommendation in text form. This recommendation guides the user on the best practices for managing and treating the identified disease, further enhancing the system's practical utility in real-world agricultural settings.

Joshi et al. (2025) trained the YOLOv8 model to detect seven tomato fruit diseases with an overall accuracy of 96.5 % mAP, 97 % precision, 93.8 % recall, and 95 % F1 score. These results are further compared with the developed system performance as presented in Table 2.

Table 2: Comparative Analysis of the Developed System and the Existing System

| S/n | Model | Accuracy % | Recall % | Precision % | F1-Score | Classes |
|-----|-------|-----------|----------|-------------|----------|---------|
| 1 | Joshi et al. (2024) | 96.50 | 97.00 | 93.8 | 95.00 | 7 |
| 2 | Developed System | 99.31 | 98.78 | 99.09 | 98.93 | 10 |
| | Improvement | 2.91 | 1.84 | 5.64 | 4.12 | |

From the result presented in Table 2, the developed system can detect ten tomato disease classes with an average accuracy of 99.31%, 98.78%, 99.09%, and 98.93% as accuracy, recall, precision, and f1-score respectively, outperformed the existing system with an improvement of 2.91%, 1.84%, 5.64%, and 4.12%. The effectiveness of the proposed approach holds the potential to improve the agricultural sector as farmers can effectively verify the diseases affecting their tomatoes by capturing leaf images using smartphones or drone technology, and the model detects the symptoms and make recommendation that can effectively improve crop yields, leading to better food security and an improved economy.

## CONCLUSION

This study has developed an improved tomato disease detection system with the ability to detect ten tomato disease classes under both controlled and uncontrolled environments. The system has an added advantage of giving recommendations on how to effectively diagnose the detected disease through a user-friendly interface with an average detection accuracy of 99.31%, 98.78%, 99.09%, and 98.93% for accuracy, recall, precision, and f1-score, respectively. This model, if adopted, can improve early tomato disease detection and offer solutions, thereby improving yield and eliminating hunger and extending of tomato species.

Corresponding author: Monday Abutu Idakwo
✉ monday.idakwo@fulokoja.edu.ng
Department of Computer Engineering, Faculty of Engineering, Federal University Lokoja, Kogi State.

*Corresponding author: Monday Abutu Idakwo*
✉ monday.idakwo@fulokoja.edu.ng
*Department of Computer Engineering, Faculty of Engineering, Federal University Lokoja, Kogi State.*